\documentclass[graybox]{svmult}

\usepackage{type1cm}        
\usepackage{makeidx}         
\usepackage{graphicx}        
\usepackage{multicol}        
\usepackage[bottom]{footmisc}

\usepackage{newtxtext}       %
\usepackage[varvw]{newtxmath}       

\renewcommand{\part}[1]{\left(#1\right)}  
\newcommand{\parg}[1]{\left\{#1\right\}}  
\newcommand{\ket}[1]{\vert {#1} \rangle}  

\newcommand{\nl}{\par\noindent}   

\newcommand{\C}{\mathbb{C}}

\makeindex             

\begin{document}

\title{Recognizing Concepts and Recognizing Musical Themes. A Quantum Semantic Analysis}
\titlerunning{Recognizing Concepts and Recognizing Musical Themes}
\author{Maria Luisa Dalla Chiara, Roberto Giuntini, Eleonora Negri, Giuseppe Sergioli}
\institute{Maria Luisa Dalla Chiara \at Dipartimento di Lettere e Filosofia, Universit\`a di Firenze. Via della Pergola 60, I-50121 Firenze, Italy; \email{dallachiara@unifi.it}
\and Roberto Giuntini \at Dipartimento di Pedagogia, Psicologia, Filosofia, Universit\`a di Cagliari. Via Is Mirrionis 1, I-09123 Cagliari, Italy; \email{giuntini@unica.it}
\and Eleonora Negri \at Scuola di Musica di Fiesole, Via delle Fontanelle 26, I-50014 Fiesole-Firenze, Italy; \email{eleonora.negri64@gmail.com}
\and {Giuseppe Sergioli } \at Dipartimento di Pedagogia, Psicologia, Filosofia, Universit\`a di Cagliari. Via Is Mirrionis 1, I-09123 Cagliari, Italy; \email{giuseppe.sergioli@gmail.com}}
%
%
\maketitle

\abstract{How are abstract concepts  and musical themes  recognized on the basis of some previous experience? It  is interesting to compare the different behaviors of human  and of  artificial intelligences with respect to this problem. Generally, a human mind that abstracts a concept (say, {\em table}) from a given set of known examples creates a {\em table-Gestalt}: a kind of vague and out of focus image that does not fully correspond to a particular table with well determined features. A similar situation arises in the case of musical themes. Can the construction of a {\em gestaltic pattern}, which is   so natural for human minds,   be taught to an intelligent machine? This problem can be successfully discussed in the framework of a quantum approach to pattern recognition and to  machine learning. The basic idea is replacing {\em classical data sets}  with {\em quantum data sets}, where either objects or  musical themes  can be formally represented as {\em pieces of quantum information},  involving the  uncertainties and the ambiguities that characterize the quantum world.  In this framework, the intuitive concept of {\em Gestalt} can be simulated by the mathematical concept of {\em positive centroid of a given quantum data set}. Accordingly,  the crucial  problem  ``how can we classify a new object  or a new musical theme (we have listened to) on the basis of a previous experience?''  can be dealt with in terms of some special {\em quantum similarity-relations}. Although recognition procedures are different for human and for artificial intelligences, there is a common method of 
``facing the problems'' that seems to work in both cases.}

\section{Introduction}
There is a story about a  king who was able to distinguish  between two different  pieces of music  only: the ``Royal March''  and the 
	``Non-Royal March''.  One is dealing with  an 
	extreme case  of a ``fully non-musical personality''. Normally people behave differently;  and 
	children  often show an early capacity of recognizing and repeating  some simple songs  they have listened to.
	Of course, such capacity generally depends on what is usually called the ``musical talent'' of each particular child. And, in the case of adult persons, one shall distinguish the behavior of professional musicians from the behavior  of generic music-listeners who may have different degrees of musical culture.\footnote{See, for instance, \cite{HH09}}

	Recognizing a musical theme is a cognitive operation that is very similar to what happens when we recognize an abstract {\em   concept} that may refer either to concrete or to ideal objects (say, {\em table, star, triangle,...}). 
	We know how quickly  children  learn to abstract concepts from the concrete objects they have met in their brief experience. On the basis of a small number of examples that have appeared in  the environment where they are living,  they easily recognize and use (mostly in a correct way) general concepts like {\em table}, {\em house}, {\em toy}. 
	
	How are abstract concepts formed and recognized on the basis of some previous experience?
	What happens in the human brain when we recognize either an abstract concept or a musical theme?   These  questions have  been intensively investigated, with different methods,  by psychologists, neuroscientists, artificial intelligence researchers, logicians, philosophers, musicians and musicologists.
	Some important  researches in the field of neurosciences (which have  used sophisticated brain-imaging techniques) have provided some partial answers. 
	Interestingly enough, neuroscientific investigations have recently interacted with an important approach to psychology: the {\em Gestalt-theory} that had been  proposed by Wertheimer, Kofka and K\"{o}hler in the early 20th century.\footnote{See, for instance, \cite{gestalt}.  }
	As is well known, a basic idea of  {\em Gestalt-psychology} is  that human perception and knowledge of objects is essentially connected with our capacity of realizing a {\em Gestalt} (a {\em form}) of the objects in question: a {\em holistic image} that cannot be identified with the set of its component elements.  The cognitive procedure goes from {\em the whole} to {\em the parts}, and not the other way around! 
	These general ideas  can be naturally applied  to investigate the question  ``how are abstract concepts formed and recognized?'' A human mind that abstracts  the concept {\em table} from a given set of concrete examples, generally creates a {\em table-Gestalt}, a kind of {\em vague and out of focus image}   that does not fully correspond to a particular table, with well determined features.  When we ask  different people the question: {\em what do you see in your mind when you hear the word  ``table''?}, we may receive different answers. An interesting answer that has been given by a person submitted to a psychological test is the following: ``I see a table, with an indefinite color, floating in an indefinite space''. In this case  the vagueness of the {\em table-Gestalt} has been expressively described by the metaphorical  image of a ``floating object''.

	Creating a {\em Gestalt} associated to a given concept is a cognitive operation that is quite natural for human intelligences. But what happens in the case of artificial intelligences? Is it possible to find  a mathematical definition for a {\em Gestalt-like concept} that  could be  taught  to an intelligent machine?  We will see  how this intriguing question can be successfully investigated in the framework of a quantum inspired  approach to  {\em pattern recognition} and to {\em machine learning}.

	From a logical point of view the relationship that connects  {\em gestaltic patterns} with particular concrete or ideal objects can be analyzed by using the concept of {\em similarity}.  Consider  a child  (let us call her {\em Alice}) who has recognized  as a {\em table} a new object that has appeared in her environment. Apparently, her recognition  is essentially based   on a quick and probably unconscious  {\em comparison}   between the main features of the new object and the ideal {\em table-Gestalt} that {\em Alice} had previously stored in her memory.
	
	Generally, any comparison involves  the use of some  {\em similarity-relations}, weak examples of relations that are 
	\begin{itemize}
		\item {\em reflexive}: any object $a$ is similar to itself;
		\item {\em symmetric}:  if $a$ is similar to $b$, then  $b$ is similar to $a$;
		\item generally {\em non-transitive}:   
		if $a$ is similar to $b$ and $b$ is similar to $c$, then $a$ is not necessarily similar to $c$.
	\end{itemize}
	Just the failure of the transitive property is one of the reasons why similarity-relations play an important role in many semantic and cognitive phenomena. A significant example  is represented by {\em metaphorical arguments}, which frequently occur in   natural languages as well in the languages of art.
	Metaphorical correlations   generally involve some allusions that are based on particular similarity-relations. Ideas  that are  currently used as possible metaphors  are often associated to concrete and visual features.  Let us think, for instance, of a visual idea that is often used as a metaphor: the image of the sea, correlated to the concepts of  immensity, of  infinity, of pleasure or fear, of places where we may get lost and die.
	
	In the tradition of scientific thought  metaphorical arguments have often been regarded as  ``fallacious''.  There is a deep logical reason that justifies such suspicion. Metaphors, based  on particular similarity-relations, do not generally {\em preserve} the properties of the objects under consideration: if  {\em Alice} is similar to {\em Beatrix} and {\em Alice} is clever, then  {\em Beatrix} is not necessarily  clever! Wrong extrapolations of properties from some objects to other  similar objects are often used in rhetoric contexts, in order to obtain a kind of {\em captatio benevolantiae}. We need only think of the soccer-metaphors that are so frequently used by many politicians!
	
	In spite of their possible ``dangers'',  metaphors have sometimes played an important role even in exact sciences.  An interesting example in logic is the current use of the
	metaphor of {\em possible world}, based on a general idea that had been deeply investigated by Leibniz. In some situations  possible worlds, that  correspond to special examples of {\em semantic models}, can be imagined as a kind 
	of  ``ideal scenes'',  where abstract objects behave as if they were playing a theatrical play. 
	And a  ``theatrical imagination'' has sometimes represented an important tool for scientific creativity, also in the search of solutions for logical puzzles and paradoxes.
	
	The classical concept of {\em possible world}  is characterized  by a strong {\em logical determinism}:  due to the {\em semantic excluded middle principle}, any sentence that refers to a given  possible world shall be either {\em true} or {\em false}.  Quantum information and quantum computation theories have recently inspired the development  of a new form of quantum-logical semantics:  classical  possible worlds  have  been replaced by   {\em vague possible worlds},  where events are generally {\em uncertain and ambiguous}, as happens in the case of microobjects.\footnote{See, for instance, \cite{Book,FSS}.}
		In the next Sections we will  see how recognition-processes that may concern either abstract concepts or musical themes can be naturally investigated in the framework of this quantum-semantic approach.

\section{A quantum semantics inspired by quantum information  theory}
		
		For the readers who are not familiar with quantum mechanics it may be useful to recall some basic concepts of the theory that play an important semantic role. 
		Suppose a physicist is studying a  quantum physical system  $\mathbf S$ (say, an electron) at a given time.
		His (her) information about   $\mathbf S$ can be  identified with a particular mathematical object that represents the
		{\em state} of $\mathbf S$ at that time. In the happiest situations our physicist might have about $\mathbf S$ 
		a {\em maximal information}  that cannot be consistently extended to a richer knowledge.  In such a case, the information in question is called a  {\em a pure state of the system}. An observer who has assigned a pure state to a given  system  knows about this system 
		all that even a hypothetical omniscient mind would know. 
		Following a happy notation introduced by  Paul Dirac, quantum pure states are  usually denoted by the expressions  $\ket{\psi},
		\ket{\varphi}, \ldots$ (where  $\ket{\ldots}$ represent  the so called  ``ket-brackets'').
		Mathematically, any pure state  $\ket{\psi}$ is a {\em vector}  (with  {\em length} $1$) living in a special abstract space, called a {\em Hilbert space} (usually indicated by the symbol $\mathcal H$).\footnote{Hilbert spaces are special examples of vector spaces that represent generalizations of geometric Euclidean  spaces. A simple example of a Hilbert space is the geometric {\em plane}, whose set of {\em points} corresponds to the set of all possible ordered pairs of real numbers.} A strange logical feature of the quantum formalism is the following: although representing a maximal piece  of information, a  pure state $\ket{\psi}$  cannot {\em decide all physical properties} that may hold for a quantum system described by $\ket{\psi}$. Due to the celebrated Heisenberg's {\em uncertainty principle} some basic properties (that may concern, for instance,  either the position or  the velocity)
		turn out to be essentially {\em indeterminate}.
		
		Generally,  a quantum pure state 
		$\ket{\psi}$ can be represented as a {\em superposition} (a vector-sum) of other pure states
		$\ket{\psi_i}$:
		
		$$ \ket{\psi}  =   \sum_ic_i\ket{\psi_i},  $$
		where $c_i$ are complex numbers (called {\em amplitudes}).
		The physical interpretation of this formal representation  is the following:
		{\em a quantum system whose state  is}
		$ \ket{\psi}$
		{\em might verify the properties   that are certain for a system in state $\ket{\psi_i}$  with a probability-value that depends on the number $c_i$.}\footnote{More precisely, this probability value is represented by the real number $|c_i|^2$ (the squared modulus of $c_i$). Since the length of $\ket{\psi}$ is $1$, we have:
			$\sum_i |c_i|^2 = 1$.
		}
		
		From an intuitive point of view one can say that a superposition-state seems to describe  a kind of  {\em ambiguous cloud of possibilities}: a set of {\em potential properties} that are, in a sense, all co-existent for a given quantum object. 
		
		Special examples of superpositions that play an important role in quantum information are represented by {\em qubits}. As is well known, in classical information theory {\em information} is measured in terms of {\em bits}. One {\em bit} represents the information-quantity that is transmitted (or received)  when one answers 
		either  ``Yes'' or  ``No''  to a given question. The two bits are usually indicated by the natural numbers  $0$ (corresponding to the answer 
		``No'' ) and 
		$1$ (corresponding to the answer 
		``Yes'' ).  On this basis, complex pieces of information are represented by sequences of many bits, called {\em registers}. For instance, a register consisting of $8$ bits  represents one {\em byte}.
		
		In quantum information theory the quantum counterpart of the classical notion of {\em bit} is the concept of {\em qubit}. In this framework, the two classical bits still exist and are represented  as two particular pure states (usually indicated by 
		$\ket{0}$ and $\ket{1}$)  that live in a special  two-dimensional Hilbert space, based on the set of all ordered pairs of complex numbers.\footnote{In this space   (usually indicated by the symbol $\C^2$)  the two classical bits
			$\ket{0}$ and $\ket{1}$ are identified with the two  number-pairs $(1,0)$ and $(0,1)$, respectively.} On this basis the concept of  {\em qubit} is then defined as any  pure state $\ket{\psi}$ (living in this space) that is a possible superposition of the two bits 
		$\ket{0}$ and $	\ket{1}$.  Thus, the typical form of a qubit is the following: 
		$$\ket{\psi} = c_0\ket{0} +   c_1\ket{1}.   $$
		
		From an intuitive point of view, the qubit 
		$\ket{\psi}$ can be interpreted as a {\em probabilistic information}: the answer (to a given question) might be  
		``No'' with a probability value that depends on the number  $c_0$ and 
		might be  
		``Yes'' with a probability value that depends on the number  $c_1$.\footnote{More precisely, the probability of the answer 
			``No''  is the number $|c_0|^2$, 
			while  the 
			probability of the answer 
			``Yes''  is the number $|c_1|^2$. }

		Not all states of  quantum systems are pure.
		More generally, a piece of 
		quantum information may correspond to a 
		{\em  non-maximal knowledge}:
		a {\em mixed state} (or {\em mixture}), that is mathematically represented as  a special   Hilbert-space operator called 
		{\em  density operator}.
		Quantum mixed states give rise to a kind of 
		{\em second degree of ambiguity}:
		while  any pure state {\em verifies with certainty} some specific quantum properties, 
		a mixed state may leave indeterminate all non-trivial quantum properties. At the same time, all pure states correspond to special examples of density operators.\footnote{Any density operator $\rho$ of a given Hilbert space can be represented  (in a non-unique way)	 as a {\em weighted sum} of some projection-operators, having the form:
			$\rho = \sum_i  w_i P_{\ket{\psi_i}}, $
			where the {\em  weights}
			$w_i$  are positive real numbers such that $\sum_i w_i = 1$,  while each
			$P_{\ket{\psi_i}}$ is the projection operator that projects over the closed subspace determined by the vector $\ket{\psi_i}$.
			Thus, any pure state  $\ket{\psi}$  corresponds to a special example of a density operator: the projection   $P_{\ket{\psi}}$.  
			The physical interpretation of  a mixed state  
			$\rho = \sum_i  w_i P_{\ket{\psi_i}} $ is the following:
			a quantum system in state $\rho$  might be in the pure state $P_{\ket{\psi_i}} $ with probability value $w_i$.}

		Complex pieces of quantum information (which may involve many qubits) are supposed to be stored by composite quantum systems (say, systems of many electrons). Thus, the quantum theoretic representation of composite systems comes into play, giving rise to one of the most mysterious features of the quantum world: {\em entanglement}, a phenomenon  that had been considered ``potentially paradoxical'' by some of the founding fathers of quantum theory (for instance,  by Einstein and by Schr\"{o}dinger).
		
		Consider a quantum composite system
		$$ \mathbf S  = \mathbf S_1 + \mathbf S_2$$
		(say, a system consisting of two electrons). \nl
		Any state of $\mathbf S$ shall live in a particular Hilbert space $\mathcal H$ that is a special product (called {\em tensor product}) of the two spaces  $\mathcal H_1$  and $\mathcal H_2$,  associated to the subsystems $\mathbf S_1$ and 
		$\mathbf S_2$, respectively. It is customary to write:
		$$ \mathcal H  = \mathcal H_1 \otimes \mathcal H_2,  $$
		where $\otimes$ indicates the tensor product.

		Unlike the case of classical physics, quantum composite systems (say, a system $\mathbf S  = \mathbf S_1 + \mathbf S_2$) have a peculiar {\em holistic} behavior: the state of the global system ($\mathbf S$)  determines the states of its component parts ($\mathbf S_1$, $\mathbf S_2$), and generally not vice versa. Thus, the procedure goes from the {\em whole} to the {\em parts}, and not the other way around. 
		Entanglement-phenomena arise in the case of particular examples of composite systems that are characterized by the following properties:
		\begin{itemize}
			\item the state of the composite system is a pure state $\ket{\psi}$ (a maximal information);
			\item this state determines the states of the parts, which  (owing to the peculiar mathematical form of $\ket{\psi}$) cannot be pure. One is dealing with {\em mixed states} that might be {\em indistinguishable} from one another.
		\end{itemize}
		
		Thus,  the information about the {\em whole} turns out to be more precise than the information about the {\em parts}. 
		Consequently, against the classical {\em compositionality-principle},
		the information about the {\em whole}  cannot be determined as a function of the pieces of information about the parts.
		Metaphorically, we might think of a strange puzzle that, once broken into its component pieces, cannot be reconstructed again, recreating its original image.

		Let us now briefly recall the basic ideas of the semantics that has been suggested by quantum information theory.\footnote{Technical details can be found in \cite{Book}.} In this semantics (which is often called {\em quantum computational semantics}) linguistic expressions (sentences, predicates, individual names,....) are supposed to denote {\em pieces of quantum information}: possible pure or mixed states of quantum systems that are storing the information in question. At the same time, logical connectives are interpreted as {\em quantum logical gates}: special 
		operators that transform the pieces of quantum information under consideration in a reversible way. Consequently, logical connectives acquire a {\em dynamic} character, representing possible computation-actions. Any {\em semantic model} of a quantum computational language assigns  to any sentence  a {\em meaning} that  lives in a Hilbert space  whose dimension depends on the linguistic complexity  of the sentence in question. 
In this way, 
 meanings  turn out to preserve, at least to a certain extent, the ``memory'' of the logical complexity of the sentences under consideration. In accordance with the quantum-theoretic formalism, quantum computational models are {\em holistic}: generally, the {\em meaning} of a compound expression (say, a sentence) determines the {\em contextual meanings}  of its well-formed parts. Thus, the procedure goes from the {\em whole} to the {\em parts}, and not the other way around, against the {\em compositionality-principle}, that had represented a basic assumption of classical semantics (strongly defended by Frege). In some interesting situations it may happen that the meaning of a sentence (say, {\em Bob loves Alice}) is an entangled pure state , while the contextual meanings of the component expressions (the names {\em Bob}, {\em Alice} and the predicate {\em loves}) are proper mixtures. In such a case the parts of our sentence turn out to be more vague and ambiguous than the  sentence itself.  One is dealing with a semantic situation that often occurs in the case of natural languages as well  in the languages of art. 
		This is one of the reasons why quantum computational semantics gives rise to some natural and interesting applications to fields that are far apart from microphysics.

\section{A quantum approach to pattern recognition and to machine learning}
		
		{\em How are abstract concepts formed and recognized on the basis of some  previous experience?} 
		This question can be successfully investigated in the framework of a 
		quantum inspired approach to pattern recognition and to machine learning,
		which  has been intensively developed in recent years.\footnote{See, for instance, \cite{Schuld}, \cite{IJ17} and \cite{Plosone}.}
		
		Consider  an {\em agent} (let us call her $Alice$) who is interested  in a given concept  $\mathcal C$ that may refer either to 
		 concrete  or to abstract objects. 
		The name $Alice$ may denote  either a  human   or 
		an artificial intelligence. 
		We will use  $Alice_H$ for a {\em human mind} and  
		$Alice_M$ for  an {\em   intelligent machine}. 
		$Alice$ will then correspond either to 
		$Alice_H$ or to $Alice_M$.

We suppose that $Alice$ (on the basis of her previous experience)  has already {\em recognized} and {\em classified} a given set of objects for which the question
``does the object under consideration verify the concept $\mathcal C$?''    
can be reasonably asked.
And we assume that the possible answers to this question are:
\begin{itemize}
	\item ``YES!''
	\item ``NO!''
	\item ``PERHAPS!''
	\end{itemize}

As an example, $Alice$ might be a child who has already recognized in the  environment where she is living:
\begin{itemize}
\item  the objects that  are tables; 
\item the objects that   are not tables. 

\end{itemize}\nl
At the same time, she might have been  doubtful about the {\em right} classification of some particular objects. For instance, she might have answered: 
``PERHAPS!''  
	to the question
``is this food-trolley a table?''.

While $Alice_H$ may have {\em seen} the objects under consideration, {\em seeing} is of course more problematic for $Alice_M$. 
Thus, generally, one shall make recourse to a  theoretic representation that faithfully describes the objects in question. 
As  happens in  physics, one can use some   convenient  mathematical objects that represent  {\em object-states}.

In the classical approach to pattern recognition and to machine learning
an {\em object-state} is usually represented  as a vector

$$ \overrightarrow{\mathbf x} = (x_1, \ldots, x_d),
$$
that belongs to the  real space $\mathbb R^d$ (the set of all ordered sequences  consisting of $d$ real numbers, where
$d \geq 1$).
Every component $x_i$ of the vector $ \overrightarrow{\mathbf x}$  is supposed to correspond to a possible value of an {\em  observable quantity}  (briefly, {\em observable})
that is considered  relevant for recognizing the concept $\mathcal C$; while  $d$ represents the number of 
the  relevant observables that are taken into consideration. 
Each  number $x_i$ is usually called a {\em  feature} of the object represented by  the vector $ \overrightarrow{\mathbf x}$. 
As an example, suppose we are referring to a class of flowers and let $\mathcal C$  correspond to a particular kind of flower (say, the {\em rose}).   We can assume that each flower-instance  is characterized by two features: the {\em petal length} and the {\em petal width}. In such a case, any object-state will be a vector
$ \overrightarrow{\mathbf x} = (x_1, x_2)$
that belongs to the space
$\mathbb R^2$.

The basic idea of  a quantum approach to pattern recognition can be sketched as follows: 
replacing  {\em classical object-states} with
{\em  pieces of quantum information}:
possible states of quantum systems that are storing the information in question. From a semantic point of view, these  {\em quantum object-states} can be regarded as possible meanings of individual names of a convenient quantum computational language.

In some pattern-recognition situations it may be useful to start with a {\em classical  information} represented by an  {\em object-state} 
$\overrightarrow{\mathbf x}.$ 
Then, the transition to a quantum pure state $\ket{\psi}$ can be realized by adopting an  {\em encoding procedure}:

$$ \overrightarrow{\mathbf x} \,\,\, \Longrightarrow
\,\,\,   \ket{\psi}_{\overrightarrow{\mathbf x} },    
$$
where  $\ket{\psi}_{\overrightarrow{\mathbf x} }$ represents the quantum pure state into which  the encoding procedure has transformed the classical object-state 
$\overrightarrow{\mathbf x}.$ 
An example of a ``natural'' encoding is the so called {\em amplitude encoding}, which is defined as follows. 

\begin{definition} Amplitude encoding \label{de:ampl}\nl
Consider an object-state
$$ \overrightarrow{\mathbf x} 
= \, (x_1, \ldots, x_d)  \in \mathbb R^d. $$
The  {\em  quantum-amplitude encoding} of
$\overrightarrow{\mathbf x}$ is the following unit vector that lives in the (real) Hilbert space $\mathbb R^{(d+1)}$:

$$ \ket{\psi}_{\overrightarrow{\mathbf x} } \, = \,  \frac {(x_1, \ldots, x_d,1)}{\lVert(x_1, \ldots, x_d,1)\rVert } $$
(where  $\lVert(x_1, \ldots, x_d,1)\rVert $ is the length of 
the vector
$(x_1, \ldots, x_d,1)$).

\end{definition}\nl
Thus, $\ket{\psi}_{\overrightarrow{\mathbf x} }$  is a   quantum pure state that preserves all  {\em   features} described by the classical object-state $\overrightarrow{\mathbf x}$.

	Of course, one could  also directly  ``reason''  in a quantum-theoretic framework, avoiding any reference to a (previously known)  classical object-state
	$\overrightarrow{\mathbf x}$. 
	In such a case, one will assume, right from the outset, that an {\em  object-state}  is represented by a  quantum pure state $\ket{\psi}$ living in a given Hilbert space.

We can  now discuss  in a quantum framework  the problem:
{\em how is a concept $\mathcal C$ recognized on the basis of a 
			previous experience?}
Suppose that  (at a given time $t_0$)  $Alice$ is interested in  the concept $\mathcal C$.  
Her previous experience concerning $\mathcal C$  can be described   by the formal notion of 
{\em   quantum} $\mathcal C$-{\em  data set} according to the following definition.

\begin{definition} Quantum $\mathcal C$-data set 
\label{de:dataset} \nl
A {\em quantum $\mathcal C$-data set} is  a sequence 
$$^\mathcal CDS  = 
\,\, (^\mathcal C\mathcal H,\,\,
^\mathcal C St, \,\, 
^\mathcal C St^+, \,\, 
^\mathcal C St^-, \,\, ^\mathcal C St^? \,\, ),    $$
where:
\begin{enumerate}
	\item[1.] $^\mathcal  C\mathcal H$ is a  finite-dimensional Hilbert space associated to $\mathcal C$.
	
	\item  [2.] $^\mathcal C St$ is a finite set of pure states  $\ket{\psi}$ of  
	$^\mathcal C\mathcal H$ for which the question
	``does the object described by $\ket{\psi}$ verify the concept $\mathcal C$?'' 
	can be reasonably asked.

	\item[3.] $^\mathcal  CSt^+$ is a  subset  of  $^\mathcal C St$ ,
	consisting of all states that have been {\em  positively classified} with respect to the concept $\mathcal C$.   
	The elements of this set are called {\em   the positive  instances} of the concept 
	$\mathcal C$.

	\item[4.] $^\mathcal  CSt^-$ is a  subset  of $^\mathcal C St$ ,
	consisting of all states that have been   {\em  negatively  classified} with respect to the concept $\mathcal C$.   
	The elements of this set are called  {\em  the negative  instances} of the concept $\mathcal C$.

	\item[5.] $^\mathcal  CSt^?$ is a (possibly empty)  subset of $^\mathcal C St$ ,
	consisting of all states that have been considered {\em  problematic} with respect to $\mathcal C$.  
	The elements of this set  are called {\em  the indeterminate instances } of the concept $\mathcal C$.

	\item[6.]  The three sets $^\mathcal  CSt^+$, $^\mathcal  CSt^-?$, $^\mathcal  CSt^?$ are pairwise disjoint.  Furthermore, 
	$^\mathcal  CSt^+ \, \cup  \, ^\mathcal  CSt^-\,
	\cup \,  ^\mathcal  CSt^? \, = \,\,
	^\mathcal  CSt.$

\end{enumerate}

\end{definition}\nl
We will indicate by 
$n^+, n^-, n^?$
the cardinal numbers of the  sets   
$^\mathcal  CSt^+$, 
$^\mathcal  CSt^-$,
$^\mathcal  CSt^?$,  respectively.

Suppose  now that  at a later time ($t_1$) $Alice$  
``meets'' a new object described by the 
object-state  $\ket{\varphi}$.
$Alice$ shall find a rule that allows her to answer the question
``does the object described by$\ket{\varphi}$ verify the concept $\mathcal C$?''
And this answer shall be based on her previous knowledge that is represented by the quantum
	$\mathcal C$-data set 
$$^\mathcal CDS  = 
\,\, (^\mathcal C\mathcal H,\,\,
^\mathcal C St, \,\, 
^\mathcal C St^+, \,\, 
^\mathcal C St^-, \,\, ^\mathcal C St^? \,\, ).    $$

A winning strategy is based on the use of two special concepts:
 {\em the quantum positive centroid} and 
		 {\em  the quantum negative  centroid}  
of a  quantum $\mathcal C$-data set.

\begin{definition}Positive and negative centroids
	\label{de:centroid} \nl
Consider a quantum $\mathcal C$-data set
$^\mathcal CDS  = 
\,\, (^\mathcal C\mathcal H,\,\,
^\mathcal C St, \,\, 
^\mathcal C St^+, \,\, 
^\mathcal C St^-, \,\, ^\mathcal C St^? \,\, ).    $

\begin{enumerate}
	
	\item[1.] The {\em  quantum positive centroid} of $^\mathcal CDS$   is the following density operator  of the space  $^\mathcal C\mathcal H$:
	
	$$\rho^+ =
	\,\, \sum_i\parg{\frac{1}{n^+}P_{\ket{\psi_i}}: 
		\, \ket{\psi_i} \in \,\,^\mathcal C St^+}. $$

	\item[2.] The {\em quantum negative centroid} of $^\mathcal CDB$   is the following density operator  of the space  $^\mathcal C\mathcal H$:
	
	$$\rho^- =
	\,\, \sum_i\parg{\frac{1}{n^-}P_{\ket{\psi_i}}: 
		\, \ket{\psi_i} \in \,\,^\mathcal C St^-}.$$

\end{enumerate}

\end{definition}

The concept of {\em  quantum positive centroid} seems to represent a  ``good'' mathematical simulation for  the intuitive idea of {\em  Gestalt}. 
Both the 
{\em quantum positive centroid} and 
the intuitive idea of {\em Gestalt} 
describe
an {\em  imaginary object}, representing a
{\em vague, ambiguous idea} that $Alice$ has obtained as an abstraction from the 
``real'' examples  she had met in her previous experience.
As happens in the case of the intuitive idea of {\em Gestalt}, the  {\em quantum positive centroid}, represented by the density operator
$\rho^+ =
\,\, \sum_i\parg{\frac{1}{n^+}P_{\ket{\psi_i}}: 
	\, \ket{\psi_i} \in \,\,^\mathcal C St^+},$
{\em  ambiguously alludes} to the concrete {\em  positive instances} that $Alice$ had previously met (which are mathematically represented by the pure states
$\ket{\psi_i} \in \,\,^\mathcal C St^+$ ).\footnote{We recall that 
	$P_{\ket{\psi_i}}$ indicates the projection operator that projects over the closed subspace determined by the vector $\ket{\psi_i}$: a special example of a density operator that corresponds to the pure state  represented by the vector 
	$\ket{\psi_i}$. According to the canonical physical interpretation of mixtures,  $\rho^+$ represents a state that ambiguously  describes a quantum system that might be in the pure state
$\ket{\psi_i}$  with probability-value 
$\frac {1}{n^+}$.}

It  is worth-while noticing that  the characteristic {\em ambiguity}  of 
quantum positive 
centroids
is not shared by  the notion of {\em  positive centroid}  that is  defined in many classical approaches to pattern recognition. 
In the classical case, a  positive centroid represents an {\em  exact object-state}, that is obtained by calculating the average values of  the values that all {\em positive instances} assign to the observables under consideration. Thus, 
unlike the quantum case, 
 classical positive centroids turn out to  describe  
{\em  imaginary objects} that are characterized by 
{\em  precise features},   without any ``cloud'' of ambiguity.

As we have noticed,  human recognitions and 
	classifications are usually performed by means of a quick and mostly unconscious {\em  comparison} between the main features  of some  new objects  we have met and a {\em  gestaltic 
		pattern} that we had  previously constructed in our mind. We also know that any
 comparison generally involves the use of some 
 {\em similarity-relations}
that are mostly grasped in a vague and intuitive way
 by  human intelligences.

 Similarity-relations play a relevant role   in the quantum theoretic formalism. 
 Important  examples of 
 {\em  quantum similarities}
 can be defined  in terms of a special function, called {\em fidelity}.
 In the case of pure states  this function is defined as follows.
 
\begin{definition} Fidelity \label{de:fidelity}\nl
Consider a Hilbert space $\mathcal H$. 
The {\em fidelity-function }	on  $\mathcal H$ is  the function $F$ that assigns to any pair   $\ket{\psi}$  and $\ket{\varphi}$  of pure states of  $\mathcal H$ the  real number 
$$F(\ket{\psi},\ket{\varphi})    = 
\arrowvert \langle \psi \mid \varphi \rangle \arrowvert^2    $$
(where $\langle \psi \mid \varphi \rangle$ is the inner product of $\ket{\psi}$  and
$\ket{\varphi}$).

\end{definition} \nl
From an intuitive point if view, the number $F(\ket{\psi},\ket{\varphi})$  can be interpreted as a measure of the the {\em degree of closeness} between the two states $\ket{\psi}$   and $\ket{\varphi}$.

The definition of {\em fidelity}  can be easily generalized to the case of density operators, which may represent either pure or mixed states.
Thus, in the general case we will write: $F(\rho, \sigma)$.

 It is interesting to recall the main properties of this  function, which  play an important role in many applications:

\begin{enumerate}

\item [1.]	$F(\rho, \sigma)    \in [0,1].$	
\item [2.] $F(\rho,\sigma)  = 
F(\sigma,\rho)  .$

\item  [3.] $F(\rho,\sigma)   = 0$  iff   
$\rho\sigma$ is the null operator.

\item  [4.] $F(\rho,\sigma)   = 1$  iff   
$\rho   = \sigma$.

	\end{enumerate}

From a physical  point of view, the fidelity-function  can be regarded as  a form of  {\em  symmetric conditional probability}: $F(\rho, \sigma) $ represents the probability that a quantum system in state $\rho$ can be transformed into a system in state $\sigma$, and vice versa.

The concept of {\em fidelity}  allows us to define in any Hilbert space $\mathcal H$ 
a special class of similarity-relations,
called 
{\em $r$-similarities}, 
	where $r$  is any real number in  the interval $[0,1]$.

\begin{definition}$r$-similarity \label{de:sim}\nl
Let  $\rho$ and $\sigma$ be two density operators of  a Hilbert space $\mathcal H$
and let $r \in [0,1]$. \nl
The state $\rho$ is called {\em  $r$-similar} to  the state $\sigma$ 
	(briefly, $\rho \not\perp_r \sigma$) iff 
	$r \le F(\rho, \sigma)$.

\end{definition}\nl
One can easily check that  (owing to the main properties of the fidelity function)  this  relation is  reflexive, symmetric and generally non-transitive.

Now $Alice_M$ has at her disposal the mathematical tools that allow her to face  
{\em  the classification-problem}.
Suppose  that $Alice_M$'s  information about a concept $\mathcal C$  is 
the  quantum $\mathcal C$-data set
$$^\mathcal CDS  = 
\,\, (^\mathcal C\mathcal H,\,\,
^\mathcal C St, \,\, 
^\mathcal C St^+, \,\, 
^\mathcal C St^-, \,\, ^\mathcal C St^? \,\, ),$$ 
\nl whose positive and  negative
centroids  are  the states $\rho^+$ and $\rho^-$, respectively.
	And let $r^*$ be 
a  {\em threshold-value} in the interval
$(\frac{1}{2}, 1]$,  that is considered relevant for $^\mathcal CDS$.
The main goal is defining  a
{\em  classifier function},
that assigns  to every state $\sigma$ (which  describes an object that $Alice_M$ may meet)

\begin{itemize}
	\item either the value $+$  (corresponding to the answer  ``YES!'');
	
	\item or the value $-$  (corresponding to the answer  ``NO!'');
	
	\item or the value $?$  (corresponding to the answer  ``PERHAPS!)''.

\end{itemize}

\begin{definition} Classifier function \label{de:classifier} \nl
Let $^\mathcal CDS  = 
\,\, (^\mathcal C\mathcal H,\,\,
^\mathcal C St, \,\, 
^\mathcal C St^+, \,\, 
^\mathcal C St^-, \,\, ^\mathcal C St^? \,\, )$  be a quantum	$\mathcal C$-data set and let $r^*$ be a threshold-value for
$^\mathcal CDS$. 	The  {\em  classifier function}
	determined by 
	$^\mathcal CDS $ and  $r^*$ is the  function
	$ {\mathcal Cl}_{[^\mathcal CDS,r^*] }$ 
	that satisfies the following condition for any state 
	$\sigma$  of   the space  $\mathcal H$:

	$ {\mathcal Cl}_{[^\mathcal CDS,r^* ]}(\sigma)  =        
	\begin{cases}  +, \,\, \text{if} \,\,\,
	\sigma  \not\perp_{r^*} \rho^+ \,\, \, \text{and  not} \,\,\, \sigma \not\perp_{r^*}  \rho^-.\\
	\text{In other words,}\,\,\, \sigma \,\,\, \text{is ``sufficiently similar''   to the positive centroid} \\ 
	\text{and is not ``sufficiently similar''  to the negative centroid.}\\
	
	-, \,\, \text{if} \,\,\,
	\sigma  \not\perp_{r^*} \rho^- \,\, \, \text{and  not} \,\,\, \sigma \not\perp_{r^*}  \rho^+.\\
	\text{In other words,}\,\,\, \sigma \,\,\, \text{is ``sufficiently similar''  to the negative centroid} \\ 
	\text{and is not ``sufficiently similar''  to the positive centroid.}\\

	?, \,\,	\text{otherwise.} \end{cases}	$
		
\end{definition}

This definition of the classifier function turns out be  quite suitable for $Alice_M$, who can now perform a simple computation in order to {\em decide} whether a new object submitted to her {\em verifies} a given concept. Apparently,
$Alice_M$'s algorithmic procedure corresponds to what is intuitively grasped by $Alice_H$, when she quickly compares her description of a new object with a {\em gestaltic pattern} stored in her memory.
Even if  recognition-procedures are  different for human and for artificial intelligences,   there is a common method of 
``facing the problems'' that seems to work in both cases.

\section{Musical themes and musical similarities}
Although
	music and quantum theory belong to two far apart worlds, the semantics suggested by quantum information theory can be successfully applied  to a formal analysis of music.\footnote{See \cite{BookMus}.}  We will see how this special form of  {\em quantum musical semantics} represents a useful tool for investigating the intriguing question of {\em musical recognitions.}

	Any musical composition  (a sonata, a symphony, a lyric opera,...)  is generally determined by three basic elements:
	
	\begin{enumerate}
		
		\item a {\em score};
		\item	a set of   {\em performances};
		\item	a set of  {\em musical ideas}  (or {\em musical thoughts}), which represent possible  {\em meanings }for systems of   {\em  musical phrases} written in the score.
		
	\end{enumerate}

	Scores represent the syntactical component of musical compositions: 
	systems of signs that are, in a sense,  similar to the  {\em formal systems} of scientific theories.
	Performances are, instead,  {\em  physical events}, that occur in space and time. 
	As is well known, not all pieces of music are associated with a score.  We need only think of folk songs  or of jazz music.  However,  in classical Western  music  compositions are usually equipped with a  score that has been written by a composer.

	Musical ideas represent a more mysterious element.
	One could ask: is it reasonable to assume the existence of such {\em ideal objects} that are similar to the {\em  intensional meanings} investigated by logic? 
	We give a positive answer to this question. In fact, 
	a musical composition cannot be simply reduced to a  score  and to a system of  sound-events. 
	Between a score   and the  sound-events created by a  performance there is something intermediate:
	the world of the  {\em musical thoughts} that underlie the different performances.
	This is the ideal world  where normally live composers and conductors, who are often accustomed to study scores without any help of a material instrument.

	The basic principle of  quantum musical semantics  is that {\em musical ideas} can be formally represented  as special cases of
	{\em pieces of quantum information}, which  may have the characteristic form of  {\em quantum superpositions}. 
	 Accordingly, we can conventionally write:
	
	$$ \ket{\mu}  =   \sum_ic_i\ket{\mu_i},  $$
	
	where:
	\begin{itemize}
		\item $\ket{\mu}$  is an abstract object representing a {\em  musical idea} that {\em  alludes} to other  {\em  ideas} $\ket{\mu_i}$ that are all co-existent:
		\item the number $c_i$ measures the 
		``importance''  of the component 
		$\ket{\mu_i}$  in the context  $\ket{\mu}$.

	\end{itemize}
	
	The use of the superposition-formalism is a powerful abstract tool that allows us to  represent, in a natural way, the {\em allusions} and the {\em ambiguities} that play an essential role in music. And  in some special cases
	musical ideas can be even represented as  peculiar {\em mixtures}, that are characterized by a deeper degree of ambiguity.	In accordance with the quantum-theoretic formalism we will use the symbols 
	$\ket{\mu}$, $\ket{\mu_1}$, $\ket{\mu_2}$, .... for {\em pure musical ideas} that behave as quantum pure states. At the same time, generic musical ideas that may behave either as pure or as mixed states will be indicated by the symbols
	$\mu$, $\mu_1$, 
	$\mu_2$, ...  .

	As is well known, an  important feature of music is the capacity of 
	evoking  some 
	{\em extra-musical meanings}:
	subjective feelings, situations that are vaguely imagined by the composer or by the interpreter or by the listener, real or virtual theatrical scenes  (which play an important role in the case of lyric operas and of {\em Lieder}).
	The interplay between {\em musical ideas} and {\em extra-musical meanings} can be naturally represented in the framework of our quantum  musical semantics:
	{\em extra-musical meanings} can be dealt with as examples of
	{\em vague possible worlds},
	where {\em  events} are generally ambiguous, as happens in the quantum world.
	
	Musical scores are characteristic 
	{\em  two-dimensional syntactical objects}
	that can be formally represented as special kinds of 
	{\em  matrices} (with {\em rows} and 
	{\em columns}). Each column contains symbols for notes or pauses that shall be performed at the same time; while each row is a sequence of  symbols corresponding to notes or pauses that shall be performed in succession. In this way, {\em chords} can be represented  as fragments of columns, while {\em melodies} can be represented as fragments of rows. The two-dimensional configuration of scores clearly reflects, in the musical notation,   the role played by parallelism in music.
	As an example  we can refer to the celebrated 
	{\em incipit } of Beethoven's Fifth Symphony (Fig.1).
	
	\begin{figure}
		\includegraphics[width=8cm]{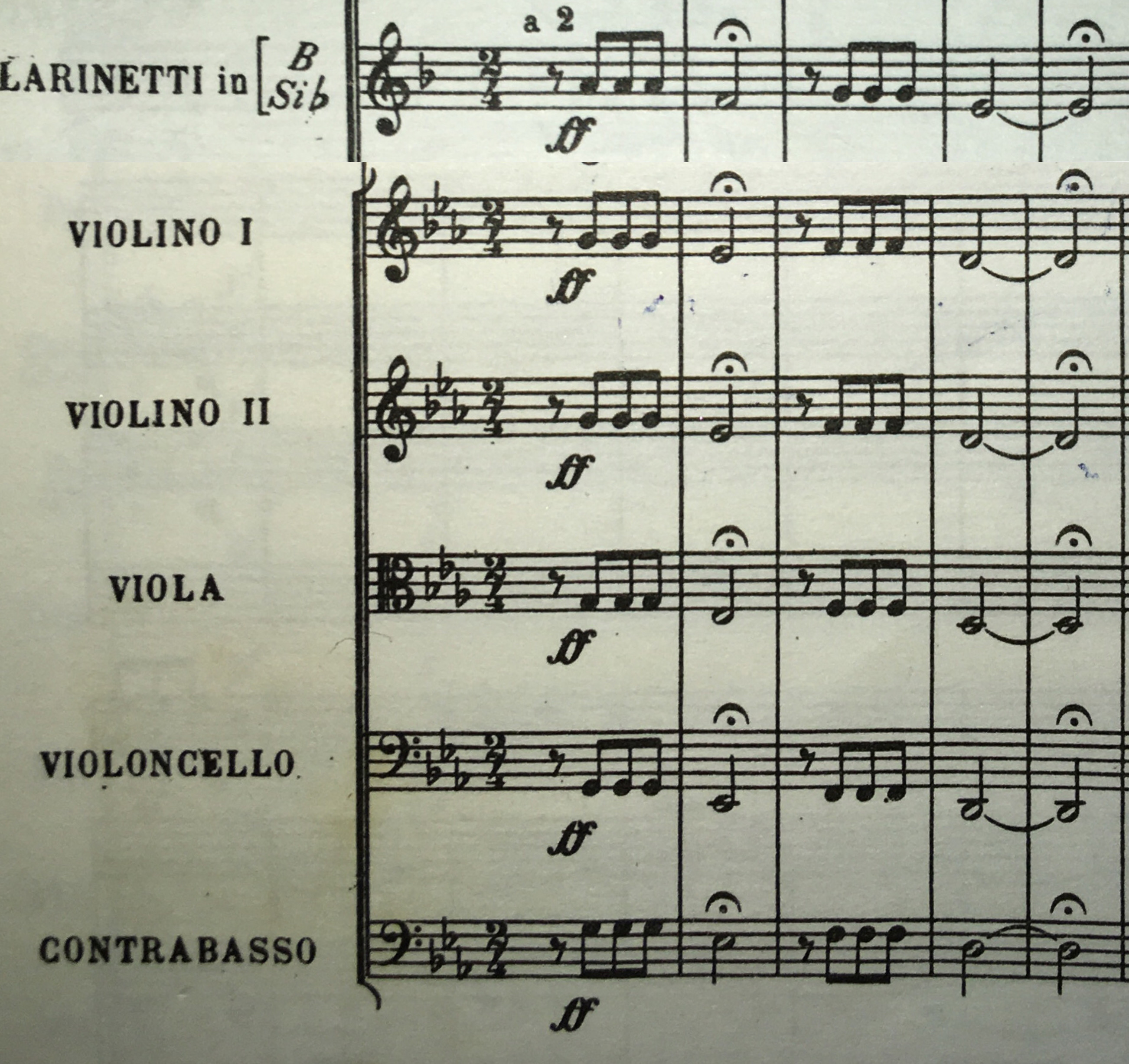}
		\caption{The incipit of Beethoven's Fifth Symphony}
	\end{figure}

	Any score is subdivided in 
	complex systems of {\em  musical phrases}. Generally, a phrase 
	may be 
	either a {\em monodic phrase}, represented by a one-dimensional {\em horizontal} fragment of the score;
	or a {\em polyphonic  phrase} (with 
	{\em  horizontal} and {\em vertical} components), represented by a two-dimensional fragment of the score.		
	Unlike the notion of  {\em sentence}  of  a formal scientific language, the concept of {\em musical phrase} does not generally represent a 
	rigid notion. 
	The subdivision of a score in {\em  musical phrases} may also depend on  the interpreter's choices.   
	And it is not by chance that  one  often  speaks of  the
	``phrasing''
	that characterizes the {\em interpretation} of a given performer.

	{\em Interpreting} a given score (say, Beethoven's Fifth Symphony) means assigning  to every 
	{\em system of musical phrases} 
	written in the score  a convenient  {\em  musical idea} that  evolves in time.
	And, as happens in the  quantum computational semantics,
	{\em  musical meanings}   have a characteristic  {\em holistic} behavior: 
	generally,  {\em the meaning  of a global phrase-system  determines the contextual meanings  of all its parts (and not the other way around)}.

	Let us consider again the incipit of
	Beethoven's  Fifth Symphony.
	And let us briefly indicate this first phrase of the symphony by
	$\mathbf{Phr^{Fifth1}}.     $
	Every {\em  interpretation} of $\mathbf{Phr^{Fifth1}}$   realizes a particular
	{\em  musical idea}.
	As is well  known, different conductors have proposed different interpretations of this famous musical phrase; and every interpretation may depend on a particular choice of the {\em dynamics} or of the {\em tempo}.
	In the framework of our quantum musical semantics  a {\em  musical idea}  that represents a possible interpretation of   the 
	phrase $\mathbf{Phr^{Fifth1}}$  can be conventionally indicated as follows:
	$$\ket{\mu^{Fifth1}}.   $$

	One could ask: what kind of abstract object is a {\em musical meaning}? Does the ket-notation, used in our quantum musical semantics, simply play a ``metaphorical''  role''?
	In fact, we could be technically more precise, representing  a musical meaning (say,  
	$\ket{\mu^{Fifth1}}$) as a {\em piece of quantum information} that, in principle, could  be stored by a quantum composite system $\mathbf S$. In this representation, each column of the score should be associated to a particular subsystem 
	of $\mathbf S$.   However, as we can imagine, the details of such a technical representation would not be interesting for the aims of a musical analysis. What is important is regarding musical meanings as special examples of {\em intensional meanings} that behave according to the general rules of the quantum  computational semantics.

	A critical question concerns 
	the relationship between {\em musical ideas} and {\em musical themes}.
	But what exactly are {\em musical themes}?
	The term ``theme'' has been used for the first time  in a musical sense by 
	Gioseffo Zarlino, in his {\em Le istitutioni harmoniche} (1558),  as a melody that is repeated and varied in the course of a musical work. 
	In the framework of our semantics the concept of  {\em  musical theme}  cannot be simply identified with a
	{\em musical idea} that represents a possible interpretation of a  particular 
	{\em  musical phrase}  (written in a given score).
	We have seen how any {\em  interpretation} of the   Fifth Symphony  associates to the first {\em phrase} of the symphony a {\em musical idea}:
	$$\ket{\mu^{Fifth1}}.   $$
	At the same time, what is usually called  {\em the main theme}  of 
	the Fifth Symphony's first movement is something  more abstract that neglects a number of {\em  musical parameters}, which may concern, for instance, the {\em pitch} or the {\em timbre}.
	Musically cultivated people generally   recognize this famous Beethoven's theme when it is played by different instruments, in different (low or high) registers  and in different (minor) tonalities.

	Which are the  characteristic  features  of this theme that represent some  {\em invariant parameters},  that  cannot be neglected?  First of all,  a particular 
	{\em   sequence of melodic intervals and pauses}. Then, a particular
	{\em   meter} ($\frac{2}{4}$) and a peculiar {\em rhythmic structure}, which is independent  of  the notes that shall be played. Thus, generally, a {\em theme} can be regarded as a highly abstract musical idea that is determined   by a sequence of melodic intervals and pauses, embedded in a given rhythmic structure.
	This suggests to consider  an abstraction from a given {\em  phrase} written in a score. 
	We can introduce the concept of  
	{\em abstract musical theme },
	which represents an {\em  invariant} with respect to possible {\em  timbre and pitch-transformations}.

	As an example, let us refer again to the first {\em  phrase} of the Fifth Symphony. 
	We will briefly represent the {\em  abstract theme} associated to this  {\em  phrase}  by  the notation described  in Fig.2.   The symbolic convention assumed in this notation is the following: 
	\begin{itemize}
		\item the squared brackets mean that we are abstracting from the ``real notes'' written in the score-fragment  inside the brackets . What we are referring to is a particular sequence  of melodic intervals and pauses, embedded in a rhythmic structure (which is determined by the score-fragment under consideration).\footnote{The use of the squared brackets is suggested by a  notation often used in mathematics, where operations  involving an abstraction are frequently indicated by the brackets $[\ldots \ldots]$.} 
		\item The ket-brackets mean that we are representing our abstract theme as a particular example of a {\em pure musical idea} $\ket{\mu}$ dealt with as a {\em meaning} in the framework of  the quantum musical  semantics.\footnote{Of course, we might use a ``more mathematical'' notation, indicating all melodic intervals by convenient arithmetical  expressions. However, this kind of  notation (which plays an important role in the framework  of computer music) would be too heavy  and hardly interesting for the aims of our semantic approach.}
	\end{itemize}
	
	\bigskip

	\bigskip
	
	\begin{figure}
		\includegraphics[width=10cm]{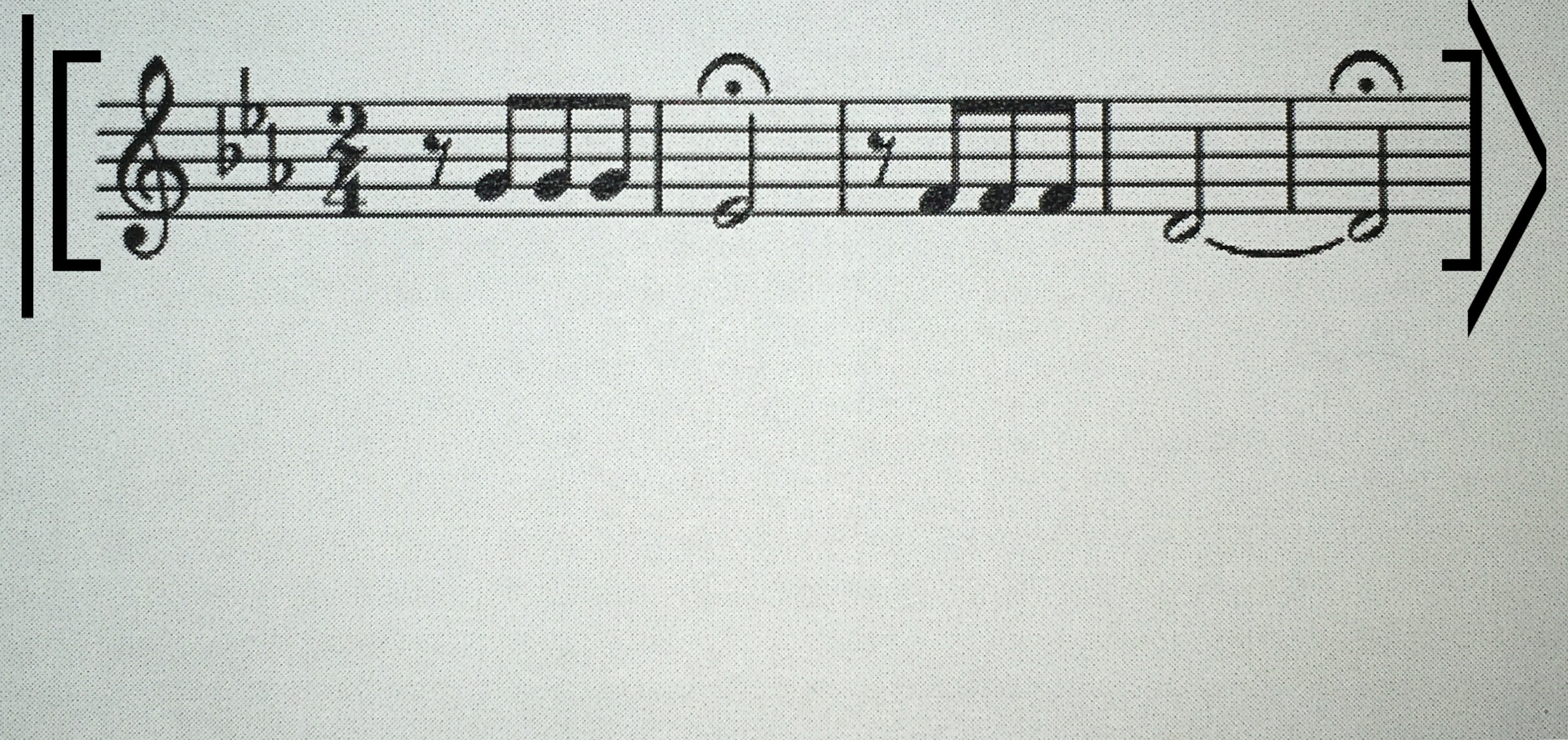}
		\caption{A notation for the main theme of the Fifth Symphony's first movement}
	\end{figure}

	We can assume that just this {\em abstract theme},
	briefly indicated by 
	$\ket{[\mu^{Fifth1}]}$,  
	can formally represent what is usually called  
	{\em  the main theme} of the Fifth Symphony's first  movement.

	Of course, {\em abstract themes } represent special examples of {\em  musical ideas}.
	On this basis we can say that: 
	the	{\em  musical idea}
	$$\ket{\mu^{Fifth1}} $$
	that represents a possible interpretation  of the 
	{\em phrase}
	$\mathbf {Phr^{Fifth1}}$
	{\em  expresses}
	the {\em abstract theme}
	$$\ket{[\mu^{Fifth1}]}.$$
	As expected, one  could  also directly associate
	{\em  abstract themes} to {\em phrases},
	asserting that 	 {\em a given phrase  expresses  a corresponding abstract theme}.
	
	So far, we have considered examples of {\em monodic abstract themes}, which correspond to fragments of {\em rows} in the formal representation of a given score. However, in some cases it may be interesting to consider also examples  of {\em polyphonic abstract themes}, that correspond to polyphonic phrases of the score in question. For instance, we might consider the complex abstract theme that corresponds to the global first phrase of the Fifth Symphony (represented in Fig. 1). In such a case, the {\em harmonic relationships} between the different parts of our theme would come  into play. Notice that what is usually called ``The Theme''  of a composition having the form {\em Theme and Variations} is  generally represented by  a  {\em polyphonic musical idea} that, in turn,  can include  a particular {\em monodic abstract theme}. And just this monodic theme represents the ``main character'' which all {\em Variations} allude to, according to modalities that may appear more or less hidden.
	
	Let us now turn to the intriguing question that concerns  musical recognitions.
		What does {\em recognizing a melody or a musical theme} mean? 
		Recognition processes in music seem to be quite similar to what happens when we recognize an {\em abstract concept}, which may refer either to concrete or to  ideal objects. 
			In the case of music, the role played  by abstract concepts   can  be naturally replaced by   {\em musical themes}. We suppose that $Alice$ is interested in a particular musical theme (say, the theme expressed by the {\em incipit} of Beethoven's 
		Fifth Symphony). The musical idea that $Alice$'s mind associates to this theme may be quite vague (depending, of course, on $Alice$'s musical preparation). What is important  is that $Alice$ can  use a {\em label} (a {\em name}) that, in principle, can refer to a particular {\em abstract musical theme}; in this case,  the theme that  we have  previously indicated  by the notation 
		$\ket{[\mu^{Fifth1}]}$.  Accordingly, in our musical applications of pattern recognition-methods we will write $\mathcal{T}$ ({\em theme}), instead of $\mathcal C$ ({\em concept}).

		As happens in the case of concepts, we  suppose that, at a given time $t_0$, $Alice$ has already classified a (finite) set of {\em pure musical ideas}
		$$MId = \parg{\ket{\mu_1}, \ldots, \ket{\mu_n }}$$
		with respect to the theme $\mathcal{T}$.
		In other words, for every musical idea $\ket{\mu_i}$ in the set  $MId$, $Alice$ has answered either ``YES!'' or ``NO!'' or  ``PERHAPS!'' to the question
		``does the musical idea $\ket{\mu_i}$ express the theme $\mathcal{T}$?'' As expected, the elements of the set $MId $ (which are formally dealt with as pieces of quantum information) represent musical thoughts (stored in $Alice$'s memory) corresponding to  pieces of music  that $Alice$  might have either listened to or  performed as a musician. 
		
		On this basis, we can naturally introduce the notion of {\em quantum musical $\mathcal T$-data set}, identified with a system
		$$ ^\mathcal TMDS\, = \, (MId,\, MId^+,\, MId^-,\, MId^?),  $$
		where:
		\begin{enumerate}
			\item [1.] $MId$ is a finite set of pure musical ideas 
			$\ket{\mu_i}$ for which  the question
			``does the musical idea $\ket{\mu_i}$ express the theme $\mathcal{T}$?'' can be reasonably asked.
			\item [2.] $ MId^+$,  $ MId^-$,  $ MId^?$ represent, respectively, the sets of  the {\em positive}, of the {\em negative} and of the {\em indeterminate  instances} for the theme $\mathcal T$.
		\end{enumerate}\nl
	We assume that any  quantum musical $\mathcal T$-data set satisfies the same conditions that we have required in the case of quantum $\mathcal C$-data sets (Definition \ref{de:dataset}).
	
	The concepts of {\em positive centroid} and of {\em  negative centroid} of a quantum musical $\mathcal T$-data set can be now naturally defined as we have done  in the case of quantum $\mathcal C$-data sets.  The  positive centroid is determined as a musical idea 
		 $\kappa^+$ that is a mixture  of all positive instances 
		 $\ket{\mu_1^+}, \, \ket{\mu_2^+},\,  \ldots,\,
		 \ket{\mu_{n^+}^+}$   of $^\mathcal TMDS$.  Each  
	 weight  occurring in $\kappa^+$ is  identified with the number 
	 $\frac{1}{n^+}$ (where $n^+$  is the number of all positive instances).
	Thus, from an intuitive point of view, $\kappa^+$  represents an {\em ambiguous musical idea} that  vaguely alludes to the  pure musical ideas 
		$$\ket{\mu_1^+}, \, \ket{\mu_2^+},\,  \ldots,\,
		\ket{\mu_{n^+}^+}.$$
		In a symmetric way, one can define the {\em negative centroid} of $^\mathcal TMDS$ as the musical idea 
		$\kappa^-$ that is a mixture of all negative instances of $INF^\mathcal T_t$. 
		
		Suppose now that at a later time $t_1$ $Alices$ listens to a new musical phrase, say the {\em incipit} of Beethoven's Fifth Symphony, which might be performed by a regular orchestra or by a piano or even simply sung by someone. And suppose that $Alice$ ask herself  ``is this piece of music the main theme of the Fifth Symphony's first movement?'' Of course, $Alice$'s answer should be based on her previous knowledge, which is formally represented by the  quantum musical $\mathcal T$-data set
		$^\mathcal TMDS$.
		
		 In such situation, our natural wish would be 
		 trying  and applying  the same {\em classifier function} that we have successfully used for concept-recognitions.  But is this  procedure possible in the case of music? 
		 We have seen how,  in the case of concepts, our  definition of the classifier function has been essentially based on the notion of $r$-similarity, which admits a precise mathematical definition in the framework of quantum information theory.  To what extent can  this strategy be reasonably extended to musical recognitions?
		 
		 It is well known that similarity-relations play a very important role in the structure of music. 
		 Musical themes  are  normally  transformed in  different ways in the framework of a given composition.  
		And all  variation-phenomena (which some authors have described as the {\em Urprinzip} of music) are characterized by the occurrence of some 
		similarity-relations. 
		We may only think of  the structure of
		 {\em Fugues},  of the {\em Sonata form } and  of the  {\em Theme and Variations-form}, where {\em abstract  themes} often appear as a kind of  ``ghosts'' in a somewhat  mysterious way. 
		
		It is customary  to distinguish different  kinds of musical similarities: 
		{\em melodic},   {\em rhythmic},  {\em harmonic}, 
		{\em timbric},....    .
		In the case of tonal music some  important  similarity-relations  are often connected with  a {\em mode-transformation}:
		from a major tonality  to a minor tonality or vice versa. 
		As  an example, let us  refer to the first movement  of Beethoven's piano sonata  op.10 n.1,  in $C$ minor (the same tonality of the Fifth Symphony).  The  primary theme of the movement   is proposed  at the very beginning (Fig. 3). One is dealing with an ascending phrase, based on the three elements of the {\em triad} of the $C$ minor key ($C$, $E$ flat, $G$).
		\begin{figure}
			\includegraphics[width=12cm]{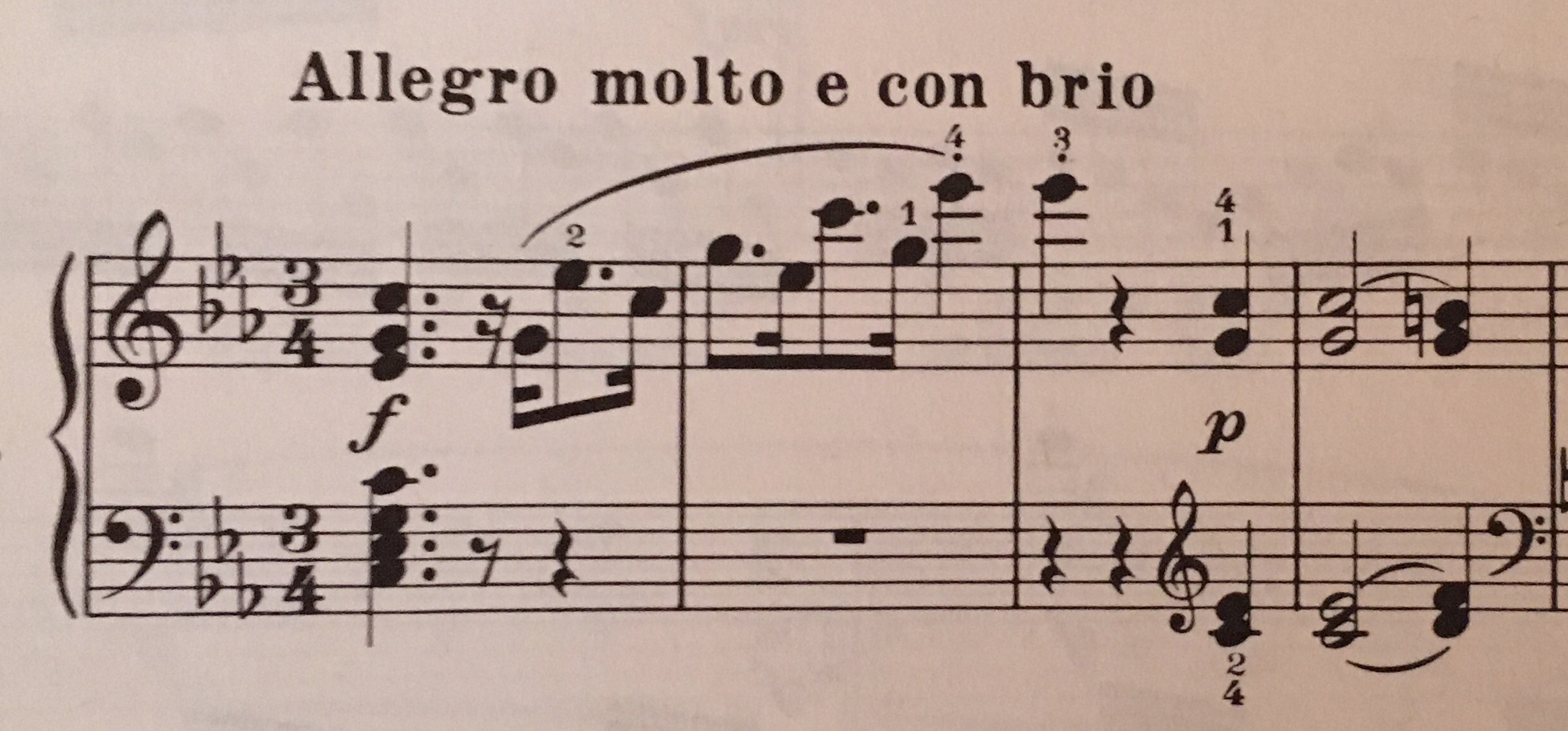}
			\caption{Beethoven, Sonata op.10, n.1. The {\em incipit} of the primary theme} 
		\end{figure}
		The dynamic indication ({\em forte}) as well as the peculiar rhythmic structure (a {\em dotted rhythm}) seem to suggest a strong statement (a kind of ``act of will''). 
		Soon after, this theme is suddenly transformed into a major mode (the $C$ major key) (Fig. 4),
		restating  (in a major version) the same strong idea that had been asserted before. 
		\begin{figure}
			\includegraphics[width=12cm]{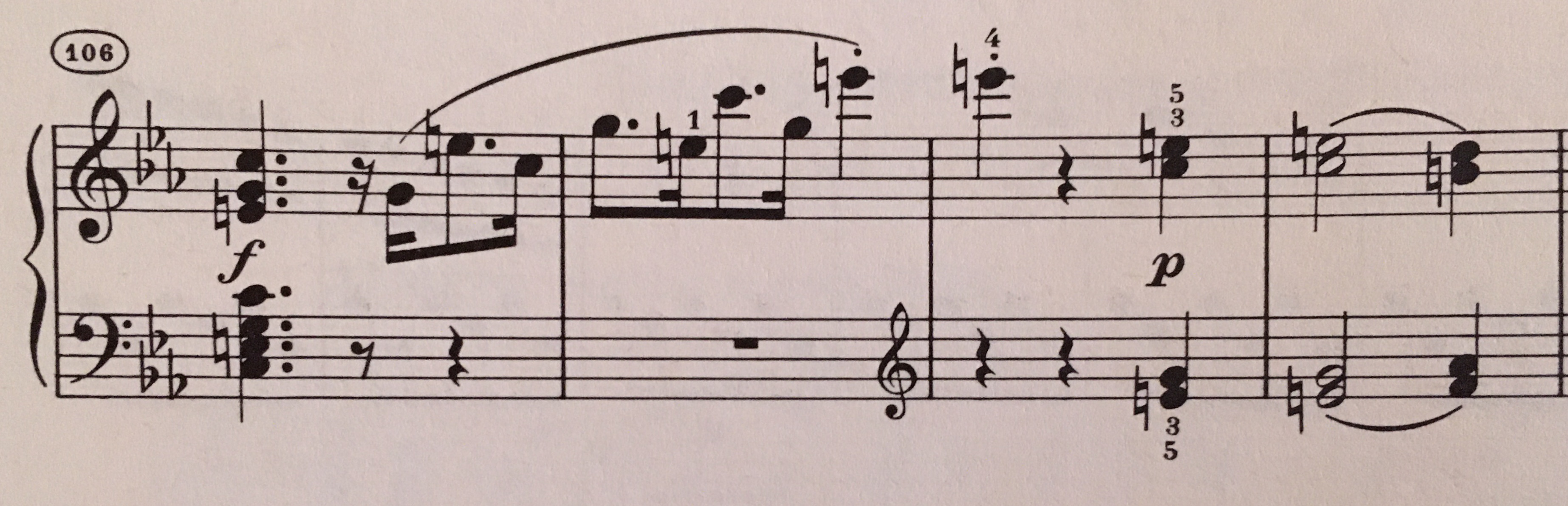}
			\caption{A major transformation  of the primary theme} 
		\end{figure} 
		
		Unlike the case of the Sonata Op.10 n.1, the main  theme of the Fifth Symphony's first movement is never directly transformed into a major  version. The phrase represented in Fig. 5 does not belong to  the symphony's score.  
		\begin{figure}
			\includegraphics[width=10cm]{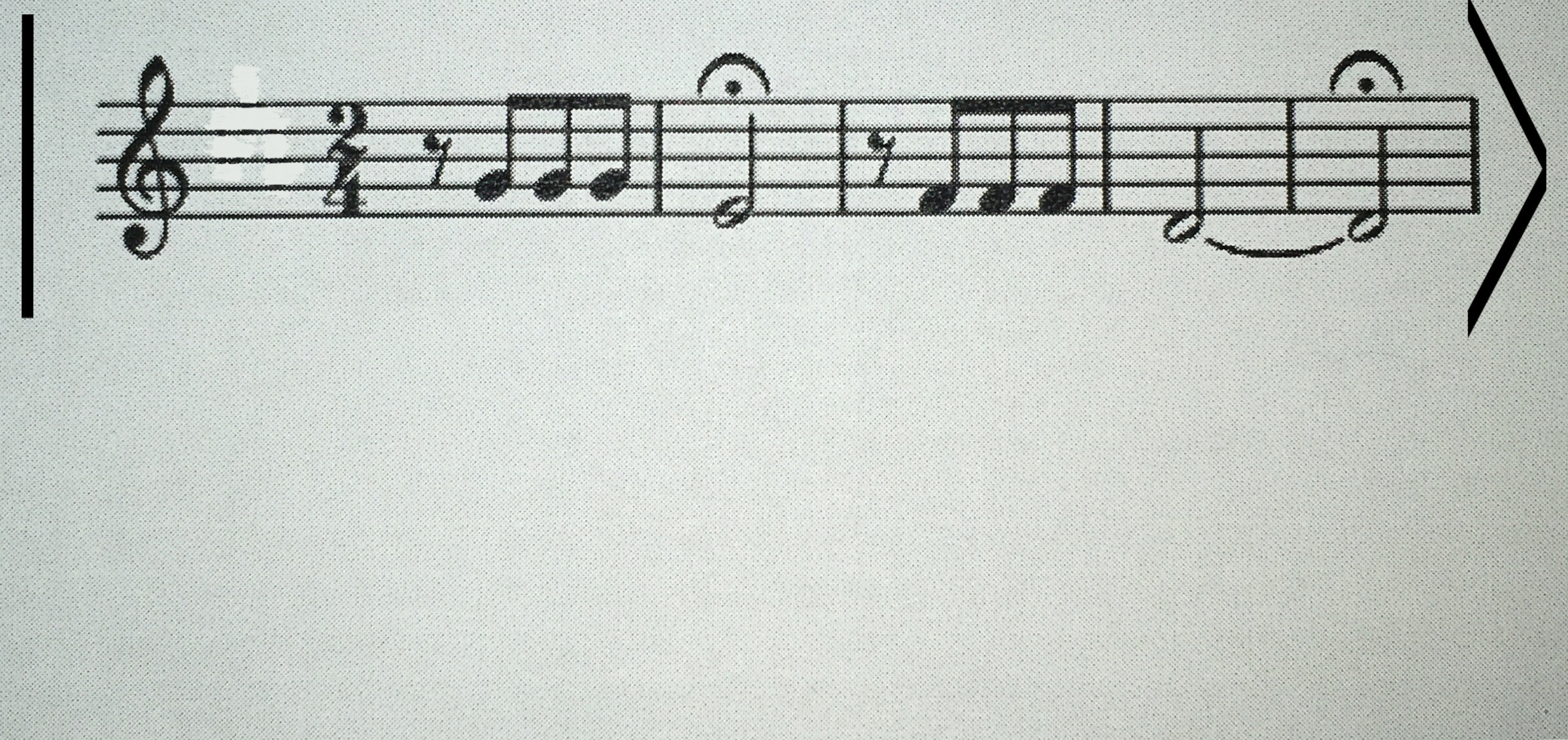}
			\caption{Beethoven, the Fifth Symphony's first movement.  A ``virtual'' major transformation  of the main theme}
		\end{figure}
		There is, however, a new theme that  appears very soon (at bars 59-62): a phrase 
		played {\em fortissimo} by the horns (in the $B$ flat major tonality)  that is naturally perceived as {\em very close} to the idea  expressed by the main theme (Fig.6).
		
		\begin{figure}
			\includegraphics[width=12cm]{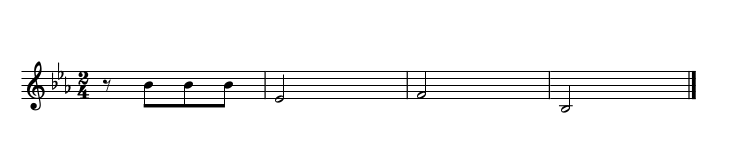}
			\caption{Beethoven, the Fifth Symphony's first movement. A major variant of the main theme}
		\end{figure}

		In the framework of our quantum musical semantics it is interesting to study  how $Alice$ lets interact  different forms of musical similarities.  For the sake of simplicity, we will now restrict our attention  to two particular  examples of similarity (which play an essential role  in the structure of music):  {\em melodic} and {\em rhythmic similarities}. Let  $\mu_1$   and
		$\mu_2 $  
		represent  two  musical ideas.  When 
		$\mu_1$  is considered {\em melodically similar} to 
		$\mu_2$, we will
		briefly write:
		$$\mu_1 \,\,Sim^M  	\,\, \mu_2 .    $$ 
		And we will write:
		$$\mu_1\,\,Sim^R  	\,\, \mu_2,   $$ 
		when $\mu_1$ is considered {\em rhythmically  similar} to  $\mu_2$.

		Melodic and rhythmic similarities can be  {\em logically combined} in different ways. In some cases it may be interesting to consider   the {\em conjunction} 
		between a melodic similarity
		$Sim^M$ and a rhythmic similarity $Sim^R$. This gives rise to a new relation, that can be called  {\em strong similarity}.  We assume (by definition) that: \nl
		{\em two musical ideas $\mu_1$ and $\mu_2$
			are strongly similar 
			if and only if they are melodically similar and rhythmically similar at the same time. }

		In some other cases it may be  interesting to consider the {\em disjunction} of the two relations 
		$Sim^M$ and $Sim^R$. This gives rise to  a different relation, that can be called  {\em weak similarity}.  
		We assume (by definition) that: \nl
		{\em  two musical ideas $\mu_1$ and $\mu_2$
			are weakly similar 
			if and only if they are either melodically similar or rhythmically similar.}
		
		As we have done in the case of concepts, we can distinguish  different
		{\em degrees of musical similarity}. 
		Although musicologists do not normally
		speak of
		{\em $r$-similarity relations}, 
		musical analyses often  use some 
		locutions  like 
		``highly similar'',
		``somewhat similar'',  ``slightly similar'', that  can be conventionally associated to some particular numerical values  (in the interval $[0,1]$).
		
		Let $Sim$ represent any form of musical similarity. When a musical idea $\mu_1$ is considered {\em $r$-similar} to a musical idea $\mu_2$, we will
		briefly write:
		$$\mu_1\,\,Sim_r  	\,\, \mu_2.    $$

		And as happens  in the case of concepts, there are    musical situations, where   it may be useful to choose a particular   {\em threshold-value} $r^*$ (in the interval $(\frac{1}{2},1]$), that is considered relevant  for the musical context under consideration. 
		Suppose, for instance,  that: 
		$\mu_1\,\,Sim_{r^*}  	\,\, \mu_2$, while $r^*$ is ``very close'' to $1$ (say, 
		$  r^* = 0.9$).    In such a case, it seems reasonable to  conclude that:
		\begin{center} {\em 
				$\mu_1$  and $\mu_1$ are highly similar}.
		\end{center}
		
		As an example, let us refer again to  the first movement of Beethoven's Sonata op.10 n.1.  Suppose that $\mu_1$ is a musical idea that corresponds to the  movement's  primary  theme  (Fig.3), while   $\mu_2$ corresponds to its major transformation (Fig.4).  Apparently, the primary theme and its major transformation have exactly the same {\em rhythmic structure}. Thus, it seems reasonable to conclude that:
		$$\mu_1 Sim^R_1   \mu_2.   $$ 
		In other words, our two  musical ideas are {\em maximally similar} from the rhythmic point of view. 
		
		The situation changes if we refer to  melodic similarity. Clearly, the primary theme and its major transformation do not have the same melodic structure, since in the major variant the minor triad ($C$, $E$ flat, $G$)  has been replaced by the major triad  ($C$, $E$, $G$).  
		However, a musical $Alice_H$, who is familiar with tonal music, ``perceives''  the two musical ideas 
		$\mu_1$  and $\mu_2$  as
		``very close'' to each other.  Hence, by a convenient choice of the threshold-value $r^*$  (for instance, by choosing $r^* = 0,9$), it seems reasonable to conclude that:
		$$ \mu_1 Sim^M_{r^*}  \mu_2. $$
		In other words,  $ \mu_1$ and $ \mu_2$  are {\em melodically very similar}.
		
		A quite different situation arises if we consider two musical ideas  $ \mu_1$  and $ \mu_2$  that correspond   to the main theme of the first movement of Beethoven's Fifth Symphony   (Fig.2)  and to its major variant (Fig.6), respectively.  In such a case, both the melodic structure  and   the rhythmic structure of 
		$ \mu_1$  and $ \mu_2$  are  different. In spite of this,  we perceive a deep  relationship that connects the musical  thoughts  expressed by 
		$ \mu_1$  and $ \mu_2$.   The major variant seems to restate,  in a more incisive and permanent way,  the strong assertion  that had been proposed by the  main theme (in a minor tonality).

		Musical similarities  are often grasped in an intuitive and rapid way by human listeners who are familiar with classical Western music. At the same time, trying to analyze, by abstract methods, the relationships  between different forms  and different degrees of similarity is not an easy task. 
		 We have seen that in the case of concepts,   $r$-similarity relations (defined in terms of the fidelity-function)  allow us to define a classifier function that has an objective and universal  behavior. This precise mathematical situation can be hardly reproduced in the case of music,
		where 
		 any particular choice  of a similarity-relation seems  to depend,  at least to a certain extent, on subjective preferences.   What we can do is referring to a class of possible musical similarity-relations, admitting that, in different contexts, we have the freedom of choosing some special elements in this class.

		 Once chosen a particular  similarity-relation  $Sim$, associated to a  threshold-value $r^*$,  it will be possible to  apply the same method used  for recognizing concepts. 
			Suppose  that $Alice$'s information  about a given abstract theme $\mathcal T$  is represented by the quantum musical $\mathcal T$-data set
		$$ ^\mathcal TMDS\, = \, (MId,\, MId^+,\, MId^-,\, MId^?),  $$
		associated to a threshold value $r^*$, and let  $\kappa^+$ and $\kappa^-$  be, respectively, the {\em positive centroid} and the {\em negative centroid}.
		As happens in the case of concepts, a   
		{\em musical classification function} $MCl$  (based on $ ^\mathcal TMDS$  and on $r^*$) shall assign to any musical idea $\nu$
		(which may represent a new musical example that  $Alice$ has listened to) either the value $+$ or the value $-$ or the value $?$. 
		
	On this basis,  we can assume by definition that:
		
		\begin{enumerate}
			\item [1.] $MCl(\nu) \,= \, + $,  
			if 
			$$ \nu \, Sim_{r^*} \,\,  \kappa^+ \,\,\,
			\text{and not} \,\,\,   
			\nu \, Sim_{r^*}\, \, \kappa^-. $$
			In other words,
			$\nu$ is {\em sufficiently  similar}  to the positive centroid  $\kappa^+$ and is not 
			{\em sufficiently similar} to the negative  centroid  $\kappa^-$. 
			
			\item [2.] $MCl(\nu) \,= \, - $,  
			if 
			$$ \nu \, Sim_{r^*} \,\,  \kappa^- \,\,\,
			\text{and not} \,\,\,   
			\nu \, Sim_{r^*}\, \, \kappa^-+ $$
			In other words,
			$\nu$ is {\em sufficiently  similar}  to the negative centroid  $\kappa^-$ and is not 
			{\em sufficiently similar} to the positive  centroid  $\kappa^+$.

			\item [3.] $MCl(\nu) \,= \, ? $,  otherwise.

		\end{enumerate}

		The application of pattern-recognition methods to musical  problems  has confirmed the interest of investigating   by abstract tools the intriguing concept of {\em musical similarity}, which has been analyzed, with different perspectives and methods, by musicians,  musicologists as well   by researchers in the field of musical informatics. 
		
		Finally, we would like to conclude with the following general remark: adopting a quantum approach to pattern recognition  has allowed us to obtain   some natural simulations for artificial intelligences   of the intrinsic  {\em vagueness and ambiguity }  that characterize many human cognitive behaviors. And, interestingly enough, one  has shown that a systematic use of  {\em quantum uncertainties} gives rise to   significant improvements of the {\em accuracy} and of the {\em algorithmic efficiency} of some machine-learning procedures.\footnote{See, for instance,
		\cite{QI18}   and \cite{Screp}.}
		These results seem to confirm that  a quantum inspired  investigation of ambiguity-phenomena can represent a powerful resource both for theoretic and for technological achievements.

\end{document}